\documentclass[letterpaper, 10pt, conference]{ieeeconf}  

\IEEEoverridecommandlockouts                              
\usepackage[caption=false,font=footnotesize]{subfig}
\usepackage{graphics} 
\usepackage{epsfig} 
\usepackage{times} 
\usepackage{amsmath} 
\usepackage{amssymb}  
\usepackage{multicol}
\usepackage{tabularx}
\usepackage[normalem]{ulem}
\usepackage{multirow}
\usepackage{tikz}
\usepackage{booktabs}
\usepackage{balance}
\usepackage{cite}
\usepackage[letterpaper, left=0.75in, right=0.75in, bottom=0.75in, top=0.75in]{geometry}

\title{\LARGE \bf
Meta Reinforcement Learning-Based Lane Change Strategy \\for Autonomous Vehicles
}

\author{Fei Ye$^{1}$, Pin Wang$^{1}$, Ching-Yao Chan$^{1}$ and Jiucai Zhang$^{2}$
\thanks{$^{1}$ F. Ye, P. Wang and C. Chan are with California PATH, University of California, Berkeley, Richmond, CA 94804, USA. 
email: {\tt\small \{fye, pin\_wang, cychan\}@berkeley.edu.}}%
\thanks{$^{2}$ J. Zhang is with GAC R\&D Center Silicon Valley Inc., 639 N. Pastoria Ave, Sunnyvale, CA 94085, USA. 
email: {\tt\small \{jzhang\}@gacrndusa.com.}}
}

\begin{document}
\maketitle
\thispagestyle{empty}
\pagestyle{empty}

\begin{abstract}
Recent advances in supervised learning and reinforcement learning have provided new opportunities to apply related methodologies to automated driving. However, there are still challenges to achieve automated driving maneuvers in dynamically changing environments. Supervised learning algorithms such as imitation learning can generalize to new environments by training on a large amount of labeled data, however, it can be often impractical or cost-prohibitive to obtain sufficient data for each new environment. Although reinforcement learning methods can mitigate this data-dependency issue by training the agent in a trial-and-error way, they still need to re-train policies from scratch when adapting to new environments. In this paper, we thus propose a meta reinforcement learning (MRL) method to improve the agent's generalization capabilities to make automated lane-changing maneuvers at different traffic environments, which are formulated as different traffic congestion levels. Specifically, we train the model at light to moderate traffic densities and test it at a new heavy traffic density condition. We use both collision rate and success rate to quantify the safety and effectiveness of the proposed model. A benchmark model is developed based on a pretraining method, which uses the same network structure and training tasks as our proposed model for fair comparison. The simulation results shows that the proposed method achieves an overall success rate up to 20\% higher than the benchmark model when it is generalized to the new environment of heavy traffic density. The collision rate is also reduced by up to 18\% than the benchmark model. Finally, the proposed model shows more stable and efficient generalization capabilities adapting to the new environment, and it can achieve 100\% successful rate and 0\% collision rate with only a few steps of gradient updates.

\end{abstract}

\section{Introduction}
Automated and semi-automated vehicles are gaining popularity in their potential use in transportation. Considerable developments have focused on autonomous driving applications during the past decade. \cite{gonzalez2015review, classical_survey, schwarting2018planning, yurtsever2019survey, talpaert2019exploring, DL_review}. Particularly noticeable in the last few years, the advancements in machine learning have prompted the application of such methodologies to the field of automated driving.

Supervised learning approaches, such as imitation learning, rely heavily on large amounts of labeled data \cite{SafeDAgger_2016}. For example, \cite{chauffeurnet} uses imitation learning to learn from human demonstrations and introduces perturbations to discourage undesirable behaviors. Each task in supervised learning is trained separately, so the trained agent is often not able to generalize to other tasks. Furthermore, acquiring and labeling a sufficient amount of data for each individual task in autonomous driving can be costly and time-consuming, and it is also challenging to cover all of the real-world driving scenarios. 

In comparison, reinforcement learning (RL) offers an alternative approach by training the agent in a trial-and-error way, and it does not require explicit human labeling or supervision on each data sample. Recently, RL methods have been increasingly applied to the decision making and control of autonomous vehicles \cite{DRL_framework_2017, DDQN-18, Pin-2018, trc_2019}. For example, the Deep Q Network (DQN) was introduced to solve high-level decision-making problems with automated speed control for highway driving \cite{DQN_itsc,DDQN-18}. A DQN model was established by Hoel et al. \cite{DDQN-18} in a simulation environment to make driving behavioral commands (e.g. change lanes to the right/left, cruise on the current lane, etc.), and the study also compare the influence of different neural network structures on the agent's performance. More recent work \cite{Tianyu-19} introduced a hierarchical architecture to train two separate policies for both high-level decision making and low-level control execution. 

Despite the rapid progress, classical reinforcement learning methods still need to re-train new policies from scratch for new tasks, which fails to exploit the learned properties from similar tasks. To truly achieve autonomous driving in the diverse and complex real-world environment, it will be essential for the autonomous driving agent to handle newly encountered situations from past experiences \cite{meta_framework_2017}. This motivates us to introduce in this work a meta reinforcement learning (MRL) approach, which integrates meta-learning into deep reinforcement learning. 

To illustrate our proposed approach, we apply the MRL method to develop a decision-making strategy for lane changing situations in highway driving. Decision-making for automated driving in a dynamically changing environment can be challenging due to the complex interactions and uncertain behaviors of other road users. With the proposed framework, the learned model can quickly adapt to new driving conditions, which may include environmental variations such as different traffic congestion levels, different road geometries, and different driving habits in different living regions. The MRL method enables the creation of a generic model that can be generalized to make automated lane-changing maneuvers for vehicles running in different traffic environments. 

Therefore, to improve the policy's generalization capability, this paper adopts a framework of model-agnostic meta learning (MAML) \cite{finn2017modelagnostic}. Beyond adapting and generalizing to new tasks more efficiently, MAML is agnostic both to the architecture of the neural network and also to the loss function. MAML is explicitly designed to train the model's initial parameters such that the model can reach the optimal performance on a new task after just a few gradient updates. Apply MAML in the context of RL also requires two loops of optimization. A meta policy parameterized by $\theta$ is updated in the outer loop by minimizing the total expected loss over all of the training tasks. The inner loop performs task-specific policy update and computes the expected loss using the updated policy. More specifically, for each task we collect trajectories using the meta initialization, and then we'll update the task-specific policy parameters by applying gradient descent to obtain the adapted policy. These trajectories collected using the adapted policy are later used to compute the meta objective, which will be sent back the outer-loop for updating the meta policy parameter $\theta$. 

Safety is an issue of paramount concern for autonomous vehicles. We also propose a novel method of tackling driving safety in conjunction with the MRL framework. In the inner loop task learner, a safety module is incorporated into the proximal policy optimization (PPO) \cite{PPO} algorithm for learning each driving task. To further enhance safety in both learning and execution phases, a safety intervention module \cite{SafeRL2017} is added to reduce the chance of taking catastrophic actions in a complex interactive environment. Moreover, we introduce a novel option of aborting lane change when the agent is making a decision of lateral action, which enables our ego-vehicle to avoid potential collisions by \emph{aborting} and changing back to the original lane at any point while undertaking the lane change action. In the longitudinal direction, the ego vehicle chooses which leading vehicle to follow, so it can perform speed adjustments even before making the actual lane change.

The rest of the paper is organized as follows: Section II describes the underlying principle of meta reinforcement learning. In section III, we  the proposed meta-reinforcement learning frameworks for learning an adaptable lane change strategy, with the detailed descriptions on establishing the reward design and model implementation. Section IV describes the simulation environment, meta-learning system architecture, state and action design in details. The simulation benchmark design, evaluation metrics, and results are presented in Section V. Section VI concludes the paper by highlighting the effectiveness of the proposed model. 
\begin{figure}[!t]
\centering
\includegraphics[width=1.0\columnwidth]{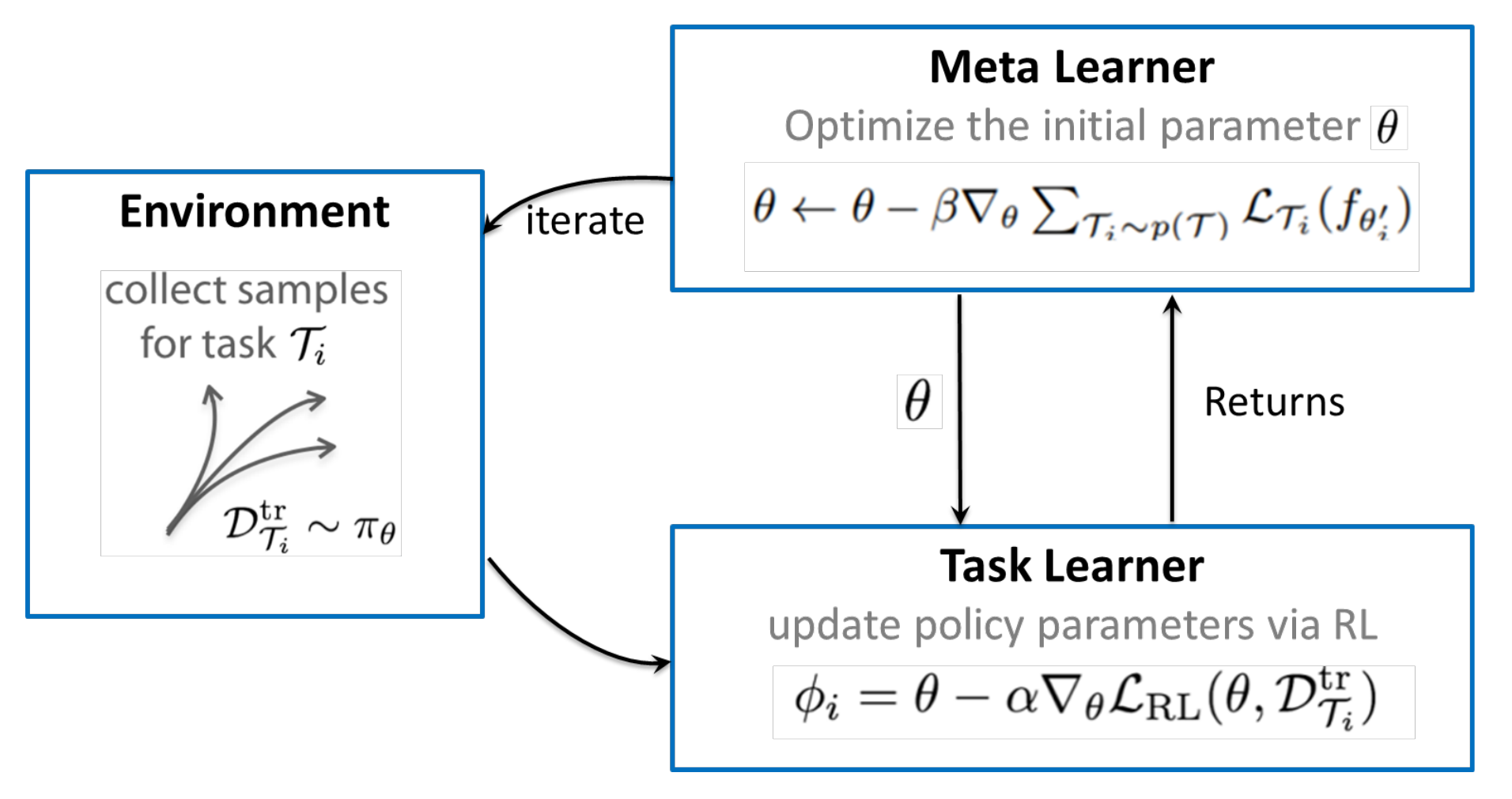}
\caption{Schematics of MAML}
\label{fig:MAML}
\end{figure}
\begin{figure*}[!t]
\centering
\includegraphics[width=0.8\textwidth]{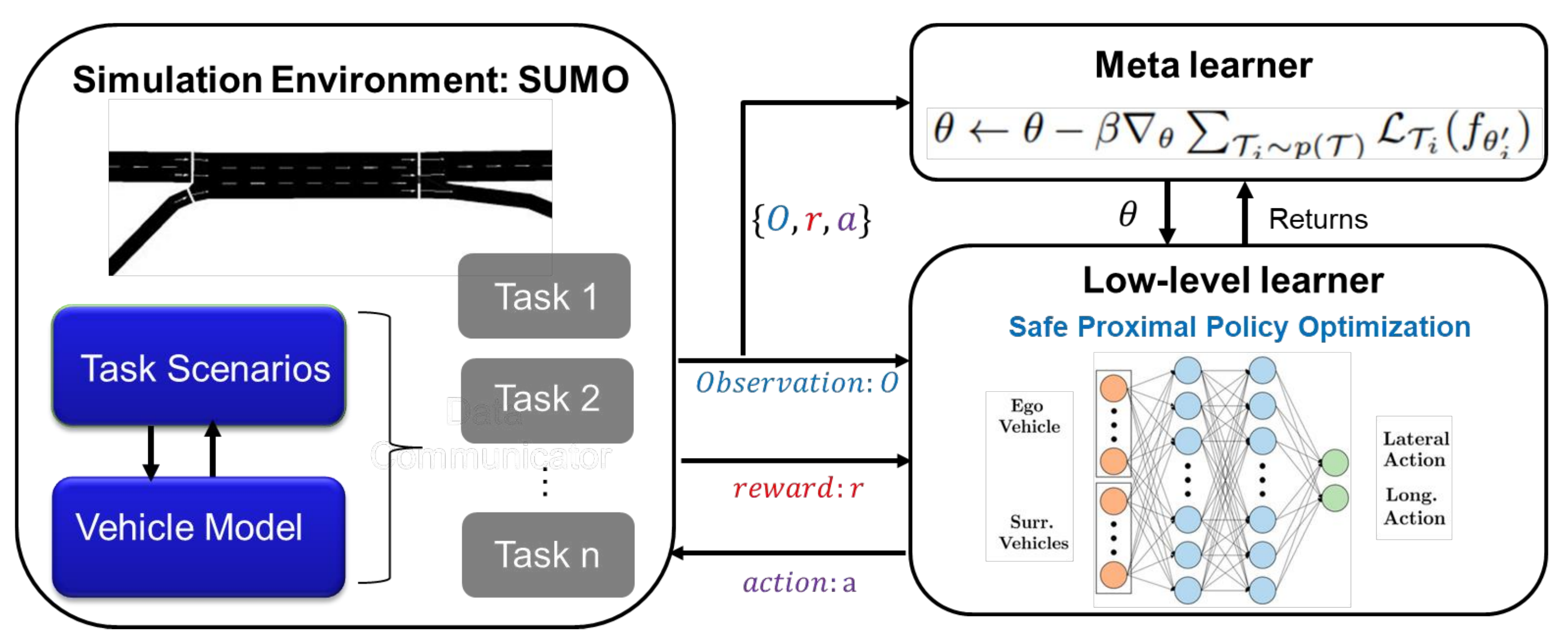}
\caption{System architecture of the proposed MAML-based lane change method}
\label{fig:framework}
\end{figure*}
\section{Meta Reinforcement Learning}
Reinforcement learning can teach how an \emph{agent} to act by interacting with its \emph{environment} in order to maximize the expected cumulative rewards for a certain task. However, reinforcement learning methods are often suffer from data-inefficiency and limited generalization. Meta learning, or learning to learn, refers to the methods can enable agents adapt quickly to new tasks using the prior knowledge or inductive biases learned from the previously seen related tasks. Recent efforts have explored the Meta-learning algorithms in the context of reinforcement learning, i.e. meta reinforcement learning \cite{finn2017modelagnostic, RL2, MRL, MRL_ICML}. The existing meta reinforcement learning algorithms can be generally categorized into two principal families. The first category of methods leverage the prior experience through the learned structure such as recurrent neural networks \cite{RL2, wang2016_MRL}.
The second set of methods is gradient-based meta reinforcement learning which attempts to optimize the parameters of the network during the meta-training such that they provide a good initialization point for adapting to new tasks with gradient descent \cite{finn2017modelagnostic, MRL}. We extend our prior work \cite{ye2020automated} that developed a policy gradient based lane change strategy for improving model generalization and task performance, and focus on the a very influential gradient-based meta reinforcement learning method which was referred to as model-agnostic meta learning (MAML) \cite{finn2017modelagnostic}.

The MAML algorithm is model-agnostic. More specifically, it is agnostic both to the architecture of the neural network and also to the loss function. The backbone of MAML is to optimize for meta parameters $\theta$ with gradient descent in two loops -- an inner loop task learner and an outer loop meta learner as shown in Fig. \ref{fig:MAML}. MAML aims to learn a good parameter initialization and use gradient descent for updating both task learner and meta learner update. These features provide many flexibilities for MAML, making it applicable to reinforcement learning problems that maximize the expected cumulative reward function through policy gradient.

Formally, we consider MAML as a neural network $f_{\theta}$ parameterized by $\theta$, which can conduct task-specific fine tuning using gradient descent when adapting to a new task. During the meta-training stage as shown in Figure \ref{fig:MAML}, MAML operates in an inner loop and an outer loop. In the inner loop, task learner initializes with the meta parameter $\theta$ and computes the updated parameter $\phi_i$ of the task learner for each task $\mathcal{T}_i$ using training data $D^{tr}_{\mathcal{T}_i}$ collected from meta-policy. And then it evaluates the loss term on the validation data $D^{vd}_i$ sampled from collected trajectories using the updated model parameters $\phi_i$. The evaluated loss for each task $\mathcal{T}_i$ can be written as
\begin{equation}
\mathcal{L}_{\mathcal{T}_{i}}\left(f_{\theta_i^{\prime}}\right) = 
\mathcal{L}\left(\phi_i, \mathcal{D}_{i}^{\mathrm{vd}}\right) = \mathcal{L}\left(\theta-\alpha \nabla_{\theta} \mathcal{L}_{RL}\left(\theta, \mathcal{D}_{i}^{\mathrm{tr}}\right), \mathcal{D}_{i}^{\mathrm{vd}}\right)
\end{equation}
where $\phi_i \leftarrow \theta- \alpha \nabla_{\theta} \mathcal{L}_{RL}\left(\theta, \mathcal{D}^{\mathrm{tr}}_i\right)$ is the updated model parameter for task $\mathcal{T}_i$.  
The loss function for updating a RL policy has the general form as
\begin{equation}
L_{RL}(\theta)=\hat{\mathbb{E}}_{t}\left[\log \pi_{\theta}\left(a_{t} | s_{t}\right) \hat{A}_{t}\right]
\end{equation}
where $\mathbb{E}_{t}$ is the expectation operator, $\pi_{\theta}$ is a stochastic RL policy, $\hat{A}_{t}$ is an estimator of the advantage function at time step $t$.
%

In the outer loop, meta learner aggregates the per-task post-update losses $\mathcal{L}\left(\phi_i, \mathcal{D}_{i}^{\mathrm{vd}}\right)$ and performs a meta-gradient update on the original model parameter $\theta$ as
\begin{equation}
    \theta \leftarrow \theta- 
\beta\cdot\nabla_{\theta}\sum_{\mathcal{T}_i \sim P(\mathcal{T})} \mathcal{L}\left(\phi_i, \mathcal{D}_{i}^{\mathrm{vd}}\right)
\end{equation}
where $\beta$ is the learning rate of the outer loop. MAML optimizes the meta-policy parameter $\theta$ such that the expected loss across all the training tasks after inner-loop update is minimized.  

At meta-test time, MAML is able to adapt the meta parameters based on a few iterations of gradient updates on the rollout trajectories collected from the new task. 
The adaptation process is represented as a few steps of gradient descent with the collected rollouts. In practice, the policy must be able to read in the state for each of the tasks, which typically requires them to at least have the same dimensionality.  

In summary, the essential idea of MAML is trying to find a good set of parameters of a neural network that does not necessarily have the optimal performance for different tasks at the meta-training stage, but can quickly adapt to new (unseen) tasks by fine-tuning with gradient descent.

%
\section{Automated Lane Change Framework Based on Meta Reinforcement Learning}
\subsection{Overview}
The framework of the proposed MAML model to perform automated lane change maneuver is illustrated in Fig. \ref{fig:framework}.
\subsubsection{Task Learner (Inner-Loop)}
The task learner learns a decision-making strategy for automated mandatory lane change maneuvers that feature safety, efficiency, and comfort. Instead of using the vanilla MAML inner-loop optimization, the REINFORCE \cite{Reinforce} algorithm as in  \cite{finn2017modelagnostic}, we further improve the learning efficiency and driving safety by implementing a proximal policy optimization (PPO) \cite{PPO} method. We further combine PPO with a safety module for inner-loop optimization. PPO algorithm is built upon an actor-critic structure, where the parameterized actor of PPO can enforce a trust region with clipped objectives, which has promising computation efficiency and learning stability. Moreover, the merit of the critic is to supply the actor with the knowledge of performance in low variance. All of these nice properties of PPO can improve its capability in real-life applications. The safety module can modify the exploration process of the task learner through the incorporation of a risk metric to further improve driving safety. 
\subsubsection{Meta-Learner (Outer-Loop)}
Meta-learner enables efficient model adaption of the lane change to new situations (i.e. different traffic density, different road geometries, different driving habits in different living regions). In the outer loop, the meta policy parameterized with $\theta$ is updated given the expected return based on trajectories collected over all the training tasks using the adapted inner-loop policy. 
\subsection{Inner-level Task-learner Design via PPO}
\subsubsection{Inner level objective}
Clipped surrogate loss objective has shown better performance when compared to the REFINFORCE loss used in the vanilla MAML algorithm. 

The clipped surrogate loss function of PPO combines the policy surrogate and a value function error term, which is defined as \cite{PPO}

\begin{equation}
\resizebox{\linewidth}{!}{
$L^{CLIP+VF+S}(\theta)=\hat{\mathbb{E}}_{t}\left[L^{CLIP}(\theta)-c_{1} L^{V F}(\theta)+c_{2} S\left[\pi_{\theta}\right]\left(s_{t}\right)\right]$}
\end{equation}
where $L^{CLIP}$ is the clipped surrogate objective, $c_1$, $c_2$ are coefficients, $L_{t}^{VF}$ is the squared-error loss of the value function $(V_{\theta}\left(s_{t}\right)-V_{t}^{\operatorname{targ}})^{2}$, and $S$ denotes the entropy loss. Specifically, the clipped surrogate objective $L_{t}^{CLIP}$ takes the following form
\begin{equation}
\resizebox{\linewidth}{!}{
$L^{CLIP}(\theta)=\hat{\mathbb{E}}_{t}\left[\min \left(r_{t}(\theta) \hat{A}_{t}, \operatorname{clip}\left(r_{t}(\theta), 1-\epsilon, 1+\epsilon\right) \hat{A}_{t}\right)\right]$}
\end{equation}
where $\epsilon$ is a hyperparameter, and $r_{t}(\theta)$ denotes the probability ratio $r_{t}(\theta)=\pi_{\theta}\left(a_{t}|s_{t}\right)/\pi_{\theta_{\text{old}}}\left(a_{t}| s_{t}\right)$. In this manner, the probability ratio $r$ is clipped at $1-\epsilon$ or $1 +\epsilon$ depending on whether the advantage is positive or negative, which forms the clipped objective after multiplying the advantage approximator $\hat{A}_{t}$. The final value of $L_{t}^{CLIP}$ takes the minimum of this clipped objective and the unclipped objective $r_{t}(\theta) \hat{A}_{t}$, which can effectively avoid taking a large policy update compared to the unclipped version \cite{PPO}, which is also known as the loss function of the conservative policy iteration algorithm \cite{CPI}.
\subsubsection{Network Structure and Optimizer}

Input layer ($1\times21$), First hidden layer ($1\times256$), Second hidden layer ($1\times256$), Output layer ($1\times6$). Adam optimizer \cite{Adam} is applied to compute and policy gradient and update the network weights. 
\subsubsection{Parameter Settings}
In terms of PPO hyper-parameters, we choose to use Adam and learning rate annealing with a step size of $1\times 10^{-4}$, and we set the horizon $T=512$, the mini-batch size as 64, the discount factor $\gamma=0.99$, and the clipped parameter $\epsilon=0.2$. More detailed simulation parameter setting is shown in Table \ref{tab:parameters}.
\begin{table}[]
\caption{Simulation parameters setup}
\centering
\resizebox{\linewidth}{!}{
\begin{tabular}{lll}
\toprule
Parameter & Value & Description                \\
\midrule
timesteps per actor batch & 512 & number of steps of environment per update                \\
clip parameter           & 0.2  & clipping range                                           \\
optim epochs             & 10    & number of training epochs per update                     \\
learning rate            & 1e-4 & learning rate                                            \\
$\gamma$    & 0.99 & discounting factor                        \\
$\lambda$   & 0.95 & advantage estimation discounting factor   \\
\bottomrule
\end{tabular}}
\label{tab:parameters}
\end{table}
\begin{table*}[]
    \caption{Near-Collision Reward Conditioned on the Action Space}
    \resizebox{\linewidth}{!}{
\begin{tabular}{lcc}
\toprule
$R_{near\_collision}$ & $C_0$: Current lane leader    $C_2$: Current lane follower  & $C_1$: Target lane leader      $C_3$: Target lane follower  \\ \midrule
Lateral action 0: Lane keeping                            & $F(C_e,C_1)$         & 0       \\
Lateral action 1: Changing to the target lane             & 0         & $\min(F(C_e,C_1), F(C_e,C_3))$       \\
Lateral action 2: Aborting lane change                    & $\min(F(C_e,C_0), F(C_e,C_2))$         & 0       \\
\bottomrule
\end{tabular}}
\label{tab:reward}
\end{table*}

\subsubsection{Reward Function}
The reward function is designed to incorporate key objectives of vehicle maneuvers, which is to develop an automated lane change strategy centered around the evaluation metrics of \emph{safety}, \emph{efficiency}, and \emph{comfort}. More specifically, these ideas are explained as follows:
\begin{enumerate}
    \item[a.] \emph{Comfort}: this reward is introduced to avoid sudden acceleration and deceleration of the vehicle that may cause vehicle occupant discomfort.
    Here, the comfort reward is an evaluation of jerks in both lateral and longitudinal direction. 
    \item[b.] \emph{Efficiency}: the ego-vehicle should manage to move to the target lane as soon as possible without exceeding the speed limit. Thus, the efficiency reward is an evaluation of relative lateral distance to the target lane, time cost and vehicle speed. 
    \item[c.] \emph{Safety}: the ego-vehicle should avoid collisions with surrounding vehicles. Thus, the safety reward is an evaluation of the risk of collisions and near-collisions. Here, we introduce near-collision penalty and a safety module to modify the exploration process by incorporation of external knowledge to further improve the driving safety during learning and execution phases. 
\end{enumerate}

The near-collision penalty $R_{near\_collision}$ and risk metric are both conditioned on the action-state space, rather than only penalizing a collision that actually takes place. In this case, an ego-vehicle $C_e$ can learn to abort the lane-change maneuver if its relative distances to the surrounding vehicles $C_i$ are smaller than a predefined threshold, indicating a collision is likely to happen. The specific form of this near-collision penalty term $R_{near\_collision}$ in terms of their relative positions is shown in TABLE \ref{tab:reward}, in which $F(C_e,C_i)$ is defined as
\begin{equation}
    F(C_e,C_i)=-1/(|P_{y\_e}-P_{y\_i}|+0.1)
\end{equation}
where $P_{y\_e}$ represents the longitudinal position of the ego-vehicle $C_e$; and $P_{y\_i}$ represents the longitudinal position of the surrounding vehicle $C_i$.

The safety module works in a reactive way that it can evaluate the collision risk and correct the learner when the chosen action causes a catastrophic result (i.e. collision with other vehicle) \cite{Shield2017}. The main idea is to distinguish the real ``catastrophic'' actions and ``sub-optimal'' actions in the automated lane changing process. The safety module can evaluate the risk of collision based on the relative distance with the surrounding vehicles in both lateral and longitudinal directions, and output a binary label to classify if the current chosen action is ``catastrophic'' or not and which vehicle is associated with the risk. Then, the safety module will
select a different safe actions based on the current state and returned label to replace the "catastrophic" action. 

More details on the formulation of reward functions that explicitly quantify the aforementioned evaluation metrics can be found in \cite{ye2020automated}.
\subsubsection{Acton Space}
We design the action space in both lateral and longitudinal directions, so that an agent can learn when and how to perform a lane change. 

In a real-world scenario, a driver’s execution of the lane change decision can also be affected by the interactions between the ego vehicle and other vehicle. Therefore, we introduce an aborting-lane-change action to the lateral action space, which enables the vehicle to abort taking a lane change action to avoid a potential collision at any point while undertaking the lane change maneuver. 

In the longitudinal direction, the ego vehicle needs to choose which leading vehicle to follow, so it can perform speed adjustments even before making the actual lane change.
\subsection{Outer-Level Meta-learner Design}
The meta-learner is an optimization-based meta-RL agent that can learn a prior parameter over a distribution of tasks that transfer via fine-tuning and gradient descent. Specifically, the process of meta-optimization in this study is updated with the following objective:
\begin{equation}
\min _{\theta} \sum_{\mathcal{T}_{i} \sim p(\mathcal{T})} \mathcal{L}_{\mathcal{T}_{i}}\left(f_{\theta_{i}^{\prime}}\right)=\sum_{\mathcal{T}_{i} \sim p(\mathcal{T})} \mathcal{L}_{\mathcal{T}_{i}}\left(f_{\theta-\alpha \nabla_{\theta}} \mathcal{L}_{\tau_{i}}\left(f_{\theta}\right)\right)
\end{equation}
6%
where $\mathcal{T}_{i} \sim p(\mathcal{T})$ represents each training task drawn from a task distribution $p(\mathcal{T})$. The model parameters $\theta$ are trained by optimizing for the performance of across all the training tasks. The network structure of the outer-level meta-learner is the same as the inner-level task-learner. Similar to the general meta-RL formulations, we assume both training and test tasks are drawn from the same task distribution $p(\mathcal{T})$, where each task is a Markov decision process (MDP), consisting of a set of actions, states, stochastic dynamics, and reward functions. 

\begin{figure}[!t]
\centering
\includegraphics[width=1.0\columnwidth]{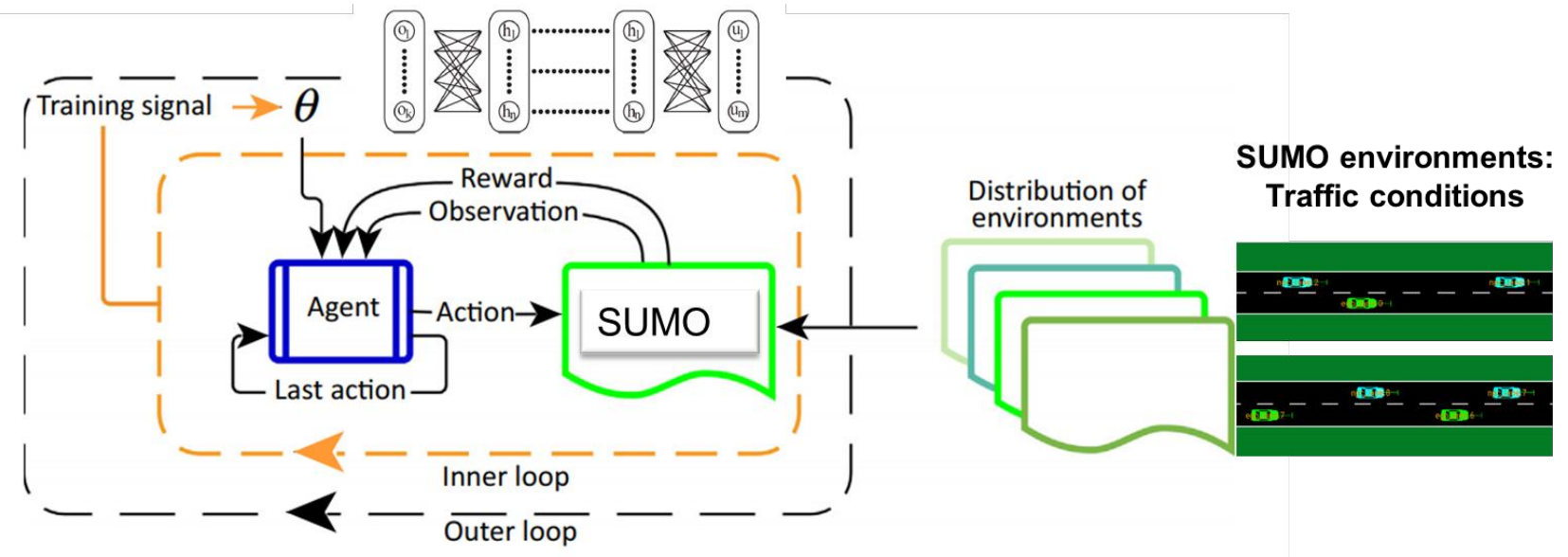}
\caption{Interactions between the simulation environment and the meta learner}
\label{fig:outer_meta}
\end{figure}

In this study, different tasks are formulated as performing lane change maneuvers at different traffic densities. At the training phase, the training tasks are drawn from those with mild to moderate traffic. At the testing phase, we evaluate our model in a new environment featuring in heavy dense traffic. An illustration of the outer-level meta-leaner interacting with different tasks of traffic conditions in the SUMO environment is illustrated in Fig. \ref{fig:outer_meta}.
\section{Simulation Experiment}
In this section, we show how we evaluate the performance of our proposed model on more challenging and unseen tasks for lane changing scenarios on highway driving, and the goal of our experiment is to demonstrate the effectiveness of the high-level meta learner combined with a low-level, safety-enhanced learner to achieve more reliable and more adaptive learning capabilities in the complex environment.
\subsection{Simulation Setup}
\begin{figure}[!t]
\centering
\subfloat[]{\includegraphics[width=3.3in]{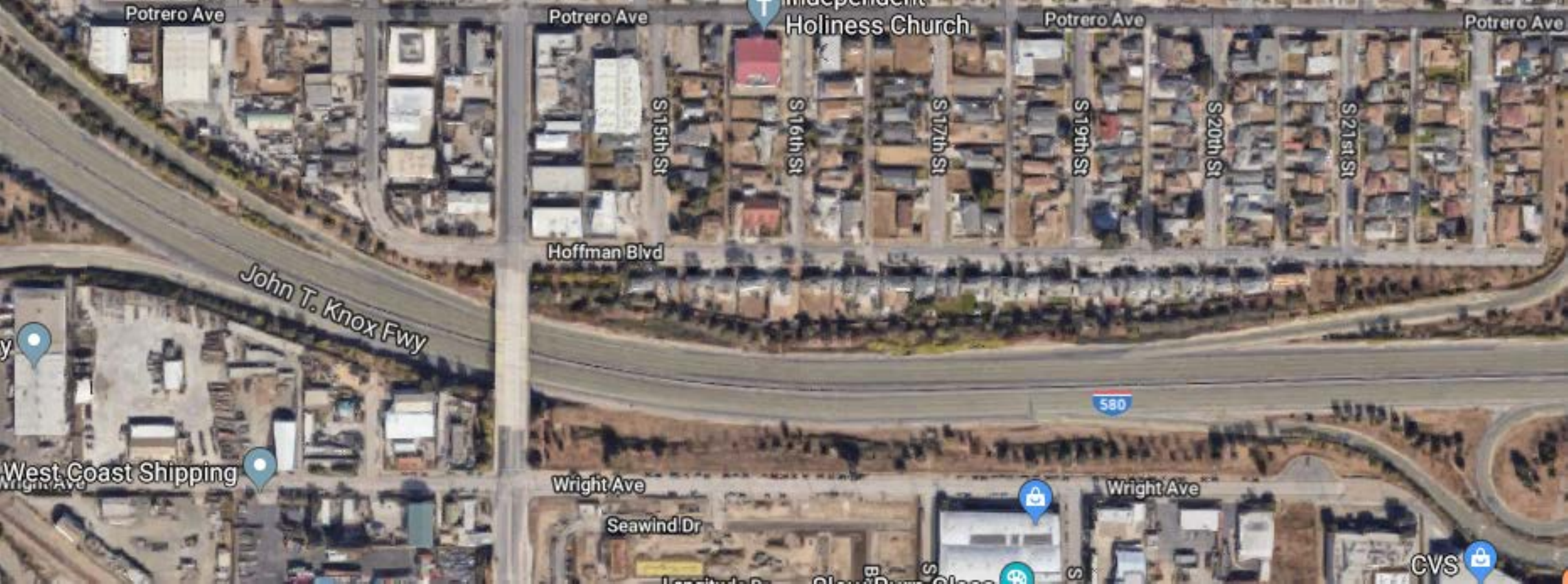}}\\
\subfloat[]{\includegraphics[width=3.3in]{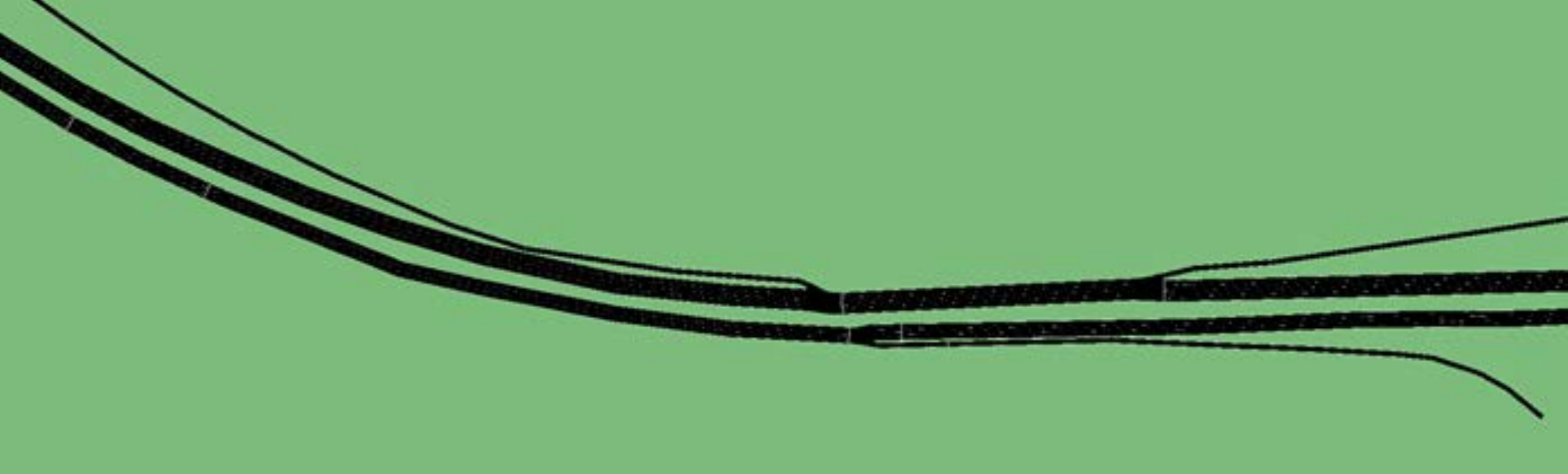}}
\caption{Simulation Network}
\label{fig:Google}
\end{figure}

The simulation network is modeled using a real-world 3-lane highway segment with on-ramps and off-ramps as shown in Fig. \ref{fig:Google}, which is implemented on SUMO \cite{SUMO}. The highway segment length to the ramp exit is 800 $m$ and width of each lane is 3.75 $m$.

The low-level vehicle control is implemented through Traffic Control Interface (TraCI). Specifically, we implemented an intelligent driver model (IDM) \cite{IDM} for car-following behavior for the other vehicles in the simulation. For the ego vehicle, we adapted the IDM for low-level longitudinal control that can adjust the vehicle speed when following different leaders. More details regarding IDM settings can be found in \cite{ye2020automated}.

In SUMO, the vehicle counts are generated from a probability factor $f\in[0,1]$, which represents the probability of a vehicle release in a second. We can simulate the highway exit behavior at light ($f\in[0,0.33)$), moderate ($f\in[0.33,0.67)$), and heavy traffic conditions ($f\in[0.67,1]$).

\begin{figure*}[!t]
\centering
\includegraphics[width=1.0\textwidth]{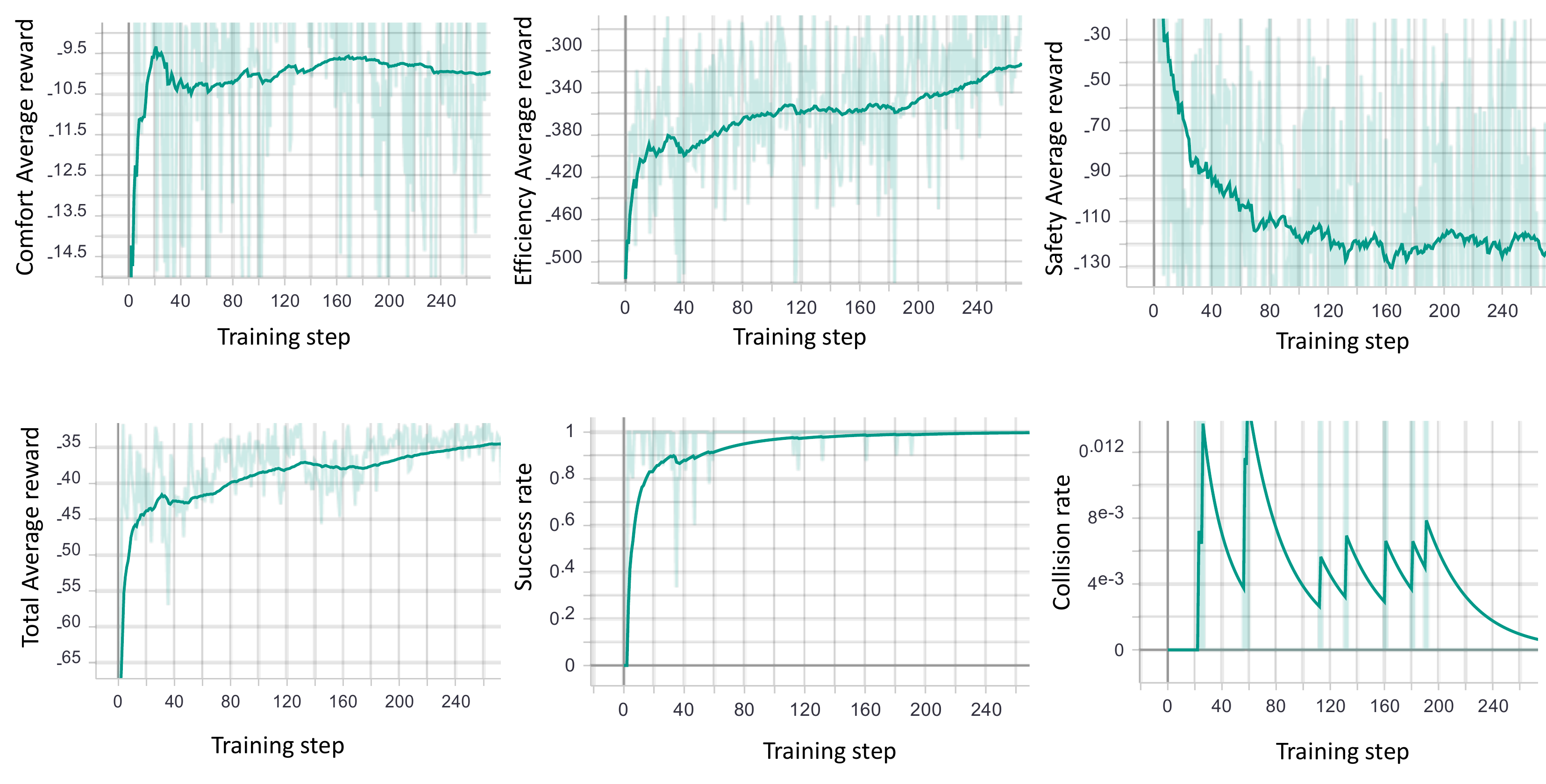}
\caption{Training results.}
\label{fig:training_results}
\end{figure*}
\subsection{Meta-RL Training and Evaluation Task Design}
\subsubsection{Meta agent}
The meta agent uses the trained meta-learner’s network weights as initialization parameters. When adapting to new tasks, a meta agent can fine-tune its policy on meta-testing tasks from meta-learner’s weights, which are obtained from the meta-learning process that involves different meta-training tasks. 

At the meta training stage, three training tasks with light to moderate traffic density (traffic probability factor $f = \{0.3,0.4,0.5\}$) will be drawn from the distribution $p(\mathcal{T})$ for meta optimization during training process. At the meta testing stage, we'll use more challenge tasks than the training tasks to evaluate the performance of the trained meta-learner. Specifically, the heavy traffic density scenario with traffic probability factor $f=0.7$ is selected as the meta testing task, and it is obvious that this task has never been seen before in the training process.
\subsubsection{Pretrained agent (Benchmark)}
To make a fair comparison, a benchmark is created based on a pretrained agent that uses the same network structure and training tasks as the meta agent. When adapting to new tasks, the pretrained agent can fine-tune its policy on meta-testing tasks based on the pretrained model, which is trained on mixed data sampled from all the training tasks.

\subsection{Evaluation Metrics}
Besides evaluating the reward function described earlier that models safety, efficiency and comfort, we also add two performance-based metrics, namely the success rate and the collision rate, to quantify the safety and effectiveness performance of the proposed MAML-based automated lane change method in both meta-training and meta-testing processes. 
\subsubsection{Success Rate} A successful task in a simulation run is defined as the ego-vehicle having successfully changed to the target lane before reaching the highway exit and avoided colliding with other vehicles. The success rate is defined as the ratio of the number of successful tasks over the total number of simulation runs in the policy rollout. The unsuccess rate can be attributes to cases of an autonomous agent failing to make a proper lane change successfully or safely before the highway exit. 
\subsubsection{Collision Rate} This metric evaluates the safety of a vehicle, as it is generally used to account for the occurrence of collision events. The collision rate is defined as the ratio of the number of collision events over the total number of simulation runs in the policy rollout.

\subsection{Results}
\begin{figure*}[!t]
\centering
\includegraphics[width=1.0\textwidth]{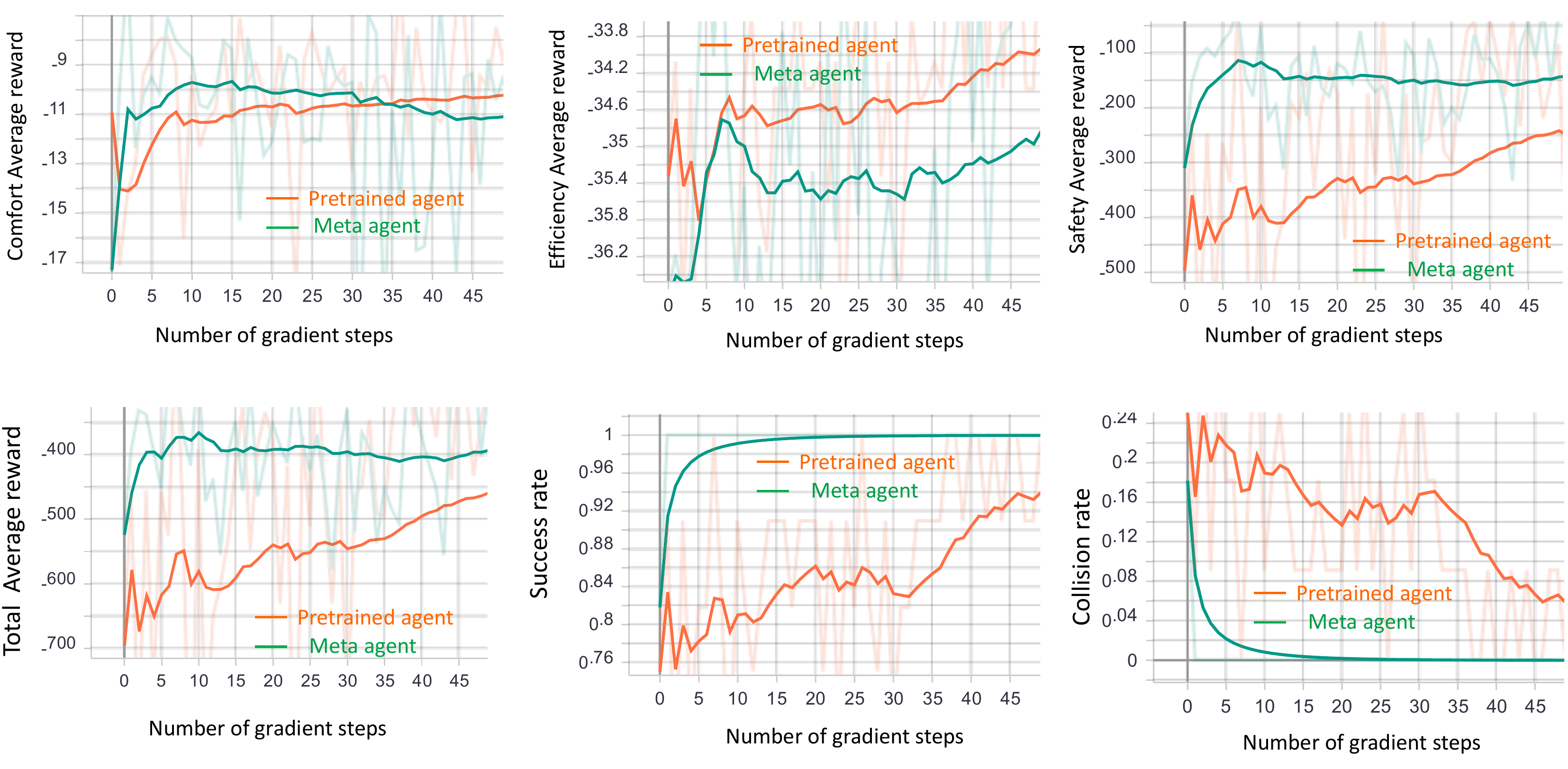}
\caption{Adaptation results.}
\label{fig:adaptation_results}
\end{figure*}
The training results of the meta agent are demonstrated in Fig. \ref{fig:training_results}, which shows the performance that the meta agent can achieve after one step gradient update in the training process. As discussed earlier, the reward function consists of three metrics: comfort evaluated by the jerk, efficiency, and safety. We can observe from Fig. \ref{fig:training_results} that both comfort and efficiency have been relatively quickly improved over the course of the training process. The safety reward slops down gradually as the training evolves. At the beginning, the meta agent has hardly made any lane change attempts at the beginning, as both success rate and collision rate are near 0. When the meta agent starts to explore by taking more lane change actions, the safety reward begins to decline. While the safety reward accounts for the occurrence of both collision and near-collision events, it can also be used to quantify the aggressiveness of lane change actions. Finally, the desired performance can be reached after performing a sufficient number of training steps, as the lane change success rate can reach $100\%$ and the collision rate can also be reduced to 0, indicating the meta agent has learned a very safe and effective lane change strategy based on the meta-training tasks. 

Fig. \ref{fig:adaptation_results} illustrated the adaptation results of both meta agent and the benchmark pretrained agent. In the meta testing stage, both agents are exposed to the same new environment, and our goal is to evaluate their adaptation capabilities in these unseen environments. Each subfigure of Fig. \ref{fig:adaptation_results} has two lines, where the orange line represents the pretrained agent, and the green one is the performance of the meta agent. Then, we'll use the same metrics to evaluate their performance first in terms of the total average return, which consists of three parts including comfort, efficiency, and safety. Then we'll compare their task performance in terms of the success rate and the collision rate.

In reinforcement learning, we typically consider the terminology of few-shot adaptation as a few-steps of gradient adaptation. As shown in Fig. \ref{fig:adaptation_results}, our meta agent is obviously performing better than the pretrained agent, or the benchmark. First of all, the meta agent achieved much better safety metrics, as its collision rate of which can quickly drop to zero after around 20 gradient steps. While both the meta agent and the pretrained agent have similar collision rate at the beginning of the meta testing phase, the pretrained agent has an $8\%$ higher collision rate in the end. Secondly, the rapid rise of the success rate up to 100\% shows our meta agent can adapt to new environments much quicker than the pretrained agent. Then, if we look at each reward metric individually, our meta agent and pretrained agent will end up with similar comfort rewards. In terms of efficiency reward, which evaluates how quickly we can switch to the desired highway exit lane, the performance of the meta agent is a bit lower than that of the pretrained agent. However, the main reason for this discrepancy is that the meta agent sacrifices efficiency for better safety, as it learns to be patient in order to avoid unnecessary collisions.

TABLES \ref{tab:success_rate} and \ref{tab:collision_rate} present the comparison between the meta agent and the pretrained agent in terms of success rate and  collision rate. It can be observed that the meta agent consistently outperforms the pretrained agent as the number of gradient steps evolves. For example, when there are only 5 gradient steps, the meta agent will have $20\%$ and $18\%$ advantage over the pretrained agent in terms of success rate and collision rate. When it reaches 20 gradient steps, the meta agent can reach $100\%$ successful rate, while the pretrained agent can only reach $86\%$. Additionally, the collision rate of the prepared agent ($14\%$) is also much greater than that of the meta agent, which is already reduced to zero.
\begin{table}[]
    \caption{Average Success Rate}
    \resizebox{\linewidth}{!}{
\begin{tabular}{lccc}
\toprule
\multirow{2}{*}{Model} & \multicolumn{3}{c}{Average Success Rate}  \\ \cmidrule{2-4} 
& $5$ gradient steps     & $20$ gradient steps   & $40$ gradient steps     \\ \midrule
Pretrained agent             & $78\%$         & $86\%$   & $90\%$    \\
Meta agent                            & $\mathbf{98\%}$         & $\mathbf{100\%}$ & $\mathbf{100\%}$       \\
Advantage                    & $20\%$         & $14\%$    & $10\%$     \\
\bottomrule
\end{tabular}}
\label{tab:success_rate}
\end{table}
\begin{table}[]
    \caption{Average Collision Rate}
    \resizebox{\linewidth}{!}{
\begin{tabular}{lccc}
\toprule
\multirow{2}{*}{Model} & \multicolumn{3}{c}{Average Collision Rate}  \\ \cmidrule{2-4} 
& $5$ gradient steps     & $20$ gradient steps   & $40$ gradient steps     \\ \midrule
Pretrained agent             & $20\%$         & $14\%$   & $8\%$    \\
Meta agent                            & $\mathbf{2\%}$         & $\mathbf{0\%}$ & $\mathbf{0\%}$       \\
Advantage                    & $18\%$         & $14\%$    & $8\%$     \\
\bottomrule
\end{tabular}}
\label{tab:collision_rate}
\end{table}
%
%
%

\section{Conclusions}
This paper proposes a strategy for automated mandatory lane change based on meta reinforcement learning. The meta agent is trained based on the MAML framework that enables the model’s quick adaptation to unseen tasks (i.e. a more challenging scenario with heavy traffic). In less than 10 steps of gradient updates, the meta agent can quickly improve the lane change performance to an average successful rate of 99\% and reduce the collision rate to less than 0.2\%. The meta agent can provide a more stable and efficient starting point for learning a new task and can achieve 100\% successful rate and 0\% collision rate with a few steps gradient updates. In summary, the meta agent consistently outperforms the pretrained agent in both adaptation speed and task performance. 
%



\balance
\bibliographystyle{IEEEtran}
\bibliography{IEEEabrv.bib,ref.bib} 
\balance

\end{document}